\documentclass{article}

\usepackage[preprint]{neurips_2024}
\usepackage{tikz}
\usetikzlibrary{
    arrows.meta,
    positioning,
    shapes,
    shapes.geometric,
    calc
}

\usepackage{graphicx}
\usepackage{amsmath}

\usepackage[utf8]{inputenc} 
\usepackage[T1]{fontenc}    
\usepackage{hyperref}       
\usepackage{url}            
\usepackage{booktabs}       
\usepackage{amsfonts}       
\usepackage{nicefrac}       
\usepackage{microtype}      
\usepackage{xcolor}         
\usepackage{titlesec}
\usepackage[most]{tcolorbox}

\titlespacing*{\paragraph}{0pt}{1mm}{4pt} 

\title{Memory as Asset: \\ From Agent-centric to Human-centric Memory Management}

\author{%
Yanqi Pan$^{1}$*, Qinghao Huang$^{2}$*, Weihao Yang$^{1}$\thanks{Co-first authors.} \\
  $^{1}$Harbin Institute of Technology, Shenzhen\\
  $^{2}$Brachisto Ltd., Hong Kong\\
  \texttt{wadepan.cs@foxmail.com, wfnuser@sjtu.edu.cn, weihao.yang00@hotmail.com} \\
}

\begin{document}

\maketitle


\begin{abstract}

  We proudly introduce Memory-as-Asset, a new memory paradigm towards human-centric artificial general intelligence (AGI). In this paper, we formally emphasize that \emph{human-centric, personal memory management} is a prerequisite for complementing the collective knowledge of existing large language models (LLMs) and extending their knowledge boundaries through self-evolution. We introduce three key features that shape the Memory-as-Asset era: (1) \emph{Memory in Hand}, which emphasizes human-centric ownership to maximize benefits to humans; (2) \emph{Memory Group}, which provides collaborative knowledge formation to avoid memory islands, and (3) \emph{Collective Memory Evolution}, which enables continuous knowledge growth to extend the boundary of knowledge towards AGI. We finally give a potential three-layer memory infrastructure to facilitate the Memory-as-Asset paradigm, with fast personal memory storage, an intelligent evolution layer, and a decentralized memory exchange network. Together, these components outline a foundational architecture in which personal memories become persistent digital assets that can be accumulated, shared, and evolved over time. We believe this paradigm provides a promising path toward scalable, human-centric AGI systems that continuously grow through the collective experiences of individuals and intelligent agents.
\end{abstract}

\begin{tcolorbox}[colback=yellow!8,colframe=black!60,title=Draft Note]
Currently, this document is an early draft describing the \emph{Memory-as-Asset} paradigm 
and its potential infrastructure for future memory systems. 
The ideas, terminology, and system design presented in this paper are still evolving. 
Some descriptions may not yet have undergone thorough proofreading or technical validation, 
and the manuscript primarily reflects our current thinking about the future 
infrastructure of memory systems. We welcome feedback and discussion from the community.
\end{tcolorbox}

\section{Introduction}

The pursuit of artificial general intelligence (AGI) fundamentally depends on the acquisition, organization, and continual evolution of knowledge. 
Recent advances in large language models (LLMs) demonstrate that large-scale collective knowledge—primarily derived from the Internet—can enable impressive general capabilities. 
However, such models largely rely on static training corpora and lack mechanisms to accumulate knowledge through persistent real-world experience. 
As intelligent systems increasingly operate as autonomous agents interacting with humans and environments, the ability to store, organize, and evolve memory over long time horizons becomes essential.

We argue that memory, rather than models alone, should be viewed as the fundamental substrate of intelligence in the near future. 
In this perspective, knowledge emerges as a function of memory: experiences recorded in memory are structured, retrieved, and synthesized to generate reasoning and decisions. 
For AGI systems, the memory space must encompass not only collective knowledge derived from global data sources, but also personal knowledge accumulated through individual interactions, as well as evolving knowledge generated through continuous exploration and reasoning.

Despite this, existing AI infrastructures primarily focus on model serving (e.g., Model-as-a-Service) and largely overlook memory as a first-class system abstraction. We have seen several efforts in memory infrastructure, but they face limitations in persistent evolution, secure ownership, and scalable sharing across agents, hindering the pursuit of AGI.

\paragraph{Local memory systems.} 
Early memory systems aim to provide persistent memory via local storage, enabling agents to accumulate experience while preserving user ownership and privacy. 
MemoryLLM~\cite{wang2024memoryllmselfupdatablelargelanguage} explores memory mechanisms built upon local large language models, enabling agents to maintain persistent knowledge through model-assisted memory retrieval and reasoning. A-MEM~\cite{xu2025mem} further organizes memories into interconnected knowledge networks using structured attributes and semantic links. 
These systems demonstrate the feasibility of persistent local memory for individuals. 
However, they operate within a single-agent scope, resulting in isolated memory silos that limit knowledge sharing and collective evolution across agents.

\paragraph{Centralized memory systems.} Recent memory systems explore memory as a service, where memory is uploaded and explicitly stored and managed on the centralized cloud. MemGPT~\cite{memgpt}, Mem0~\cite{mem0}, and MemOS~\cite{li2025memos_long} introduce structured memory management frameworks but rely on centralized cloud infrastructures, where memory is owned and controlled by platforms rather than individuals. 
Such designs violate an important principle for long-term intelligent agents: personal experiences and knowledge should remain under the ownership of the human user. 
Furthermore, centralized memory services often create isolated memory silos across agents, limiting collaborative knowledge accumulation.

\paragraph{Decentralized memory systems.}
Recent work has begun exploring decentralized mechanisms for collaborative memory evolution. 
For example, EvoMap~\cite{EvoMap_evolver} extends the OpenClaw memory architecture by enabling decentralized knowledge propagation across agents. 
Through interaction-driven memory evolution, agents can collectively share and refine knowledge rather than maintain fully isolated memory spaces. 
However, such approaches typically assume that agents automatically contribute their memories to shared evolutionary processes, thereby weakening the ownership and permission controls required for human-centric memory systems. 

Our in-depth analysis shows that the root cause of these limitations lies in the fact that existing approaches are built from an \emph{agent-centric} viewpoint, where memory is treated as an internal component serving agent execution and system optimization. Under this perspective, memory primarily functions as a system-level resource—used for retrieval, reasoning, or task completion—rather than as a persistent knowledge asset owned by humans, which inherently involves privacy, ownership, and long-term control over personal knowledge.

However, ``human-centric'' is a fundamental premise for developing AGI~\cite{asimov1942runaround}. Despite the rapid progress of large-scale models trained on Internet-scale data, a significant portion of valuable knowledge comes from human interactions, experiences, and accumulated contextual understanding. 
As agents increasingly act as digital extensions of human users, their memories inevitably record personal experiences, preferences, and knowledge generated through human–AI interaction. 
Consequently, memory should not be regarded merely as an auxiliary capability of agents, but rather as a human-centric asset that represents the long-term accumulation of individual knowledge.

In this work, we propose a new paradigm called \textbf{Memory as Asset}, which treats memory as a first-class digital asset owned and controlled by individuals. 
We refer to this property as \textbf{Memory in Hand}, emphasizing that memory remains under the direct ownership of the human user rather than centralized platforms. 
Under this paradigm, memory becomes a persistent and composable knowledge resource that can be securely shared, exchanged, and evolved across agents.

The Memory-as-Asset paradigm is characterized by three key principles: (1) \emph{Memory in Hand (Human-Centric Ownership).}
Memory represents accumulated personal experiences and knowledge, and should therefore remain under the ownership and control of individuals. 
This principle requires decentralized memory management mechanisms that ensure portability, privacy, and long-term ownership. (2) \emph{Memory Group (Collaborative Knowledge Formation).}
Personal memories should not exist as isolated \emph{memory islands}. 
Instead, memory objects can form dynamic \emph{memory groups} through permission-controlled sharing across agents and humans. 
Within such groups, memories become composable knowledge assets that can be selectively accessed, reused, or exchanged. (3) \emph{Memory Evolution (Continuous and Collective Knowledge Growth).}
Through controlled interaction among memory groups, agents can augment their personal memory spaces with knowledge obtained from others. 
This process enables continuous experience-driven learning, allowing personal memories to evolve over time while preserving ownership boundaries.

Together, these principles establish a decentralized memory ecosystem in which individuals retain ownership of their knowledge while benefiting from collaborative intelligence across agents.

We envision that the future of memory systems will shift from agent-centric architectures to \emph{human-centric ecosystems}, where humans, agents, and applications continuously create, exchange, and evolve memory assets. 
Such an ecosystem enables persistent, human-centric, and collaborative intelligence, laying the foundation for AGI with long-term human–AI coevolution.
\section{Related Work}

\subsection{LLM in the Wild: Agents With No Memory}

Early large language model (LLM) agents primarily relied on prompt-based reasoning and tool invocation without persistent memory. 
A representative framework is ReAct~\cite{yao2023react}, which interleaves reasoning and action by prompting language models to generate intermediate reasoning traces before interacting with external tools or environments. 
Subsequent systems such as AutoGPT and BabyAGI extended this paradigm to autonomous agent frameworks that decompose tasks and invoke external tools.

However, these systems largely operate in a stateless manner, where the agent's behavior depends only on the current prompt and the limited context window. 
Knowledge acquired during interactions cannot be persistently accumulated or reused across tasks. 
As a result, agents cannot develop consistent preferences, learn from past experiences, or evolve their strategies over time, highlighting the need for persistent memory mechanisms.

\subsection{Agents With Local Memory}

To address the limitations of stateless agents, recent work has introduced persistent memory mechanisms for LLM-based agents through local memory architectures. 
These systems allow agents to accumulate experiences and maintain knowledge across interactions while preserving user ownership and privacy.

MemoryLLM~\cite{wang2024memoryllmselfupdatablelargelanguage} explores memory mechanisms built upon local large language models, enabling agents to maintain persistent knowledge through model-assisted memory retrieval and reasoning. 
A-MEM~\cite{xu2025mem} further organizes memories into interconnected knowledge networks using structured attributes and semantic links to support long-term knowledge organization.

These systems demonstrate that persistent local memory can improve agent consistency, contextual awareness, and experience accumulation. 
However, most existing approaches treat memory as an application-level component tightly coupled with individual agents. 
As a result, memory remains confined within a single-agent scope, forming isolated memory silos that prevent knowledge sharing and collaborative evolution across agents.

\subsection{Centralized Memory Systems}

More recent work explores memory as a shared infrastructure for AI agents. 
Systems such as MemGPT~\cite{memgpt}, Mem0~\cite{mem0}, MemOS~\cite{li2025memos_long}, and EverMemOS~\cite{evermemos} provide centralized memory services that allow agents to store, retrieve, and update memories through unified interfaces.

These systems typically rely on centralized cloud infrastructures and backend storage systems such as vector databases or knowledge graphs to support semantic retrieval and contextual recall. 
By externalizing memory management, they allow multiple agents to access a shared memory backend.

However, centralized memory services introduce two fundamental limitations. 
First, memory ownership is controlled by the platform rather than the human user, which conflicts with the principle that personal knowledge and experiences should remain under individual ownership. 
Second, these systems primarily treat memory as a storage resource rather than an evolving knowledge asset.

\subsection{Decentralized Memory Systems}

Recent work has begun exploring decentralized mechanisms for collaborative memory evolution. 
For example, EvoMap~\cite{EvoMap_evolver} extends the OpenClaw memory architecture to enable decentralized knowledge propagation across agents. 
Through interaction-driven evolution, agents can refine and expand knowledge collectively rather than maintaining fully isolated memory spaces.

However, such approaches often assume that agents automatically contribute their memories to shared evolutionary processes, which weakens the ownership and permission controls required for human-centric memory systems.

\subsection{Summary}
These observations reveal a fundamental tension in existing memory systems: systems that preserve personal ownership lack mechanisms for collaboration, while systems that enable collaborative knowledge evolution often weaken human control over memory assets. 
This motivates the need for a new paradigm that simultaneously supports personal ownership, controlled sharing, and collaborative memory evolution.
\section{Motivation: Human-Centric Memory for Knowledge Evolution}

A central challenge in achieving Artificial General Intelligence (AGI) is enabling systems to continuously acquire, organize, and evolve knowledge while ensuring that the resulting intelligence remains aligned with human interests. 
Although modern large language models (LLMs) demonstrate strong reasoning and generation capabilities, their knowledge is largely static, derived from pretraining corpora, and lacks mechanisms for persistent experience accumulation and alignment with individual users.

From a human-centric perspective, the goal of AGI is not merely to expand knowledge, but to expand \emph{beneficial knowledge} while suppressing harmful or misaligned knowledge. 
Therefore, the design of intelligent systems must explicitly consider how memory systems accumulate, filter, and evolve knowledge in ways that serve human users.

\subsection{Memory and Knowledge Representation}\label{subsec:motivation:repre}

We first formalize the relationship between memory and knowledge.

Let $M$ denote the \textbf{memory space}, which contains all stored experiences, observations, and structured representations accumulated by an intelligent system. Memory may include interaction logs, task outcomes, learned abstractions, and knowledge derived from both pretraining and runtime experience.

Let $K$ denote the \textbf{knowledge space}, representing the actionable knowledge that an agent can utilize for reasoning and decision-making.

We define a knowledge extraction function:

\begin{equation}
K = F(M)
\end{equation}

where $F(\cdot)$ represents the process of retrieving, organizing, and synthesizing knowledge from memory. This process may include retrieval, summarization, abstraction, and reasoning.

However, not all extracted knowledge is beneficial. Some knowledge may be misleading, unsafe, or misaligned with human interests. 
We therefore partition the knowledge space into two subsets:

\begin{equation}
K = K^{+} \cup K^{-}
\end{equation}

where

\begin{itemize}
\item $K^{+}$ denotes \textbf{beneficial knowledge}, which contributes positively to human goals and well-being.
\item $K^{-}$ denotes \textbf{harmful knowledge}, which may lead to unsafe behaviors, misinformation, or actions misaligned with human interests.
\end{itemize}

From a human-centric perspective, AGI systems should maximize the accumulation and utilization of $K^{+}$ while minimizing the influence of $K^{-}$.

\subsection{Types of Knowledge and Memory Representation}

To support human-centric intelligence, an AGI system must integrate knowledge from multiple sources. 
We categorize knowledge into three classes and explicitly map them to their corresponding memory representations.

\begin{enumerate}

\item \textbf{Collective Knowledge ($K_c$)}

Knowledge derived from large-scale public data sources such as the internet, books, and training corpora used for LLM pretraining. 
This knowledge represents the collective knowledge accumulated by humanity.

We denote the corresponding memory space as \textbf{collective memory} $M_c$. 
In practice, $M_c$ is typically embodied in pretrained model parameters or shared knowledge repositories.

\item \textbf{Personal Knowledge ($K_p$)}

Knowledge accumulated through direct interaction between agents and individual users. 
This includes personal preferences, contextual information, task histories, and experiences generated through human–AI interaction.

This knowledge is stored in \textbf{personal memory} $M_p$, which records user-specific experiences and interaction histories.

\item \textbf{Evolving Knowledge ($K_e$)}

New knowledge that emerges through reasoning, synthesis, and exploration based on both collective and personal knowledge.

We denote the corresponding storage as \textbf{evolving memory} $M_e$, which contains newly synthesized knowledge structures produced during system operation.

\end{enumerate}

The overall memory space can therefore be expressed as

\begin{equation}
M = M_c \cup M_p \cup M_e
\end{equation}

Knowledge in each category is dynamically extracted from its corresponding memory:

\begin{equation}
K_i = F(M_i), \quad i \in \{c,p,e\}
\end{equation}

where $F(\cdot)$ denotes the knowledge extraction function defined in \S\ref{subsec:motivation:repre}.

\subsection{Human-Centric Knowledge Filtering}

Since the knowledge extracted from memory may contain both beneficial and harmful information, AGI systems require mechanisms to ensure that knowledge remains aligned with human interests.

We therefore introduce a human-centric filtering operator applied during knowledge extraction:

\begin{equation}
K_{human} = \Phi(F(M))
\end{equation}

where $\Phi(\cdot)$ represents a knowledge alignment process that filters harmful knowledge and retains beneficial knowledge.

As defined earlier, the extracted knowledge space can be partitioned as

\begin{equation}
K = K^{+} \cup K^{-}
\end{equation}

where $K^{+}$ represents beneficial knowledge and $K^{-}$ represents harmful knowledge.

The filtering operator ensures that only beneficial knowledge is used for reasoning and decision-making:

\begin{equation}
K_{human} = K^{+}
\end{equation}

This filtering process plays a critical role in maintaining the human-centric property of evolving intelligent systems.

\subsection{Memory Evolution}

A key capability required for AGI is the ability to continuously evolve memory through interaction and knowledge synthesis.

Among the three memory types, collective memory $M_c$ is largely static, as it originates from pretrained models or external knowledge sources. 
In contrast, personal memory $M_p$ and evolving memory $M_e$ grow dynamically during system operation.

In particular, evolving memory captures newly synthesized knowledge derived from both collective knowledge and accumulated personal experiences.

We model the evolution of evolving memory as

\begin{equation}
M_{e,t+1} = E(M_{p,t} \parallel M_{e,t})
\end{equation}

where

\begin{itemize}

\item $M_{p,t}$ denotes the personal memory accumulated from user interactions up to time $t$,

\item $M_{e,t}$ denotes the evolving memory generated through previous reasoning and synthesis processes,

\item $\parallel$ represents the combination of the two memory sources,

\item $E(\cdot)$ denotes the memory evolution operator that performs reasoning, abstraction, and knowledge synthesis.

\end{itemize}

Through this process, new evolving memory structures are generated and stored in $M_e$, enabling the continuous growth of system knowledge.

The resulting evolving knowledge can then be extracted as

\begin{equation}
K_e = F(M_e)
\end{equation}

This formulation highlights that knowledge evolution in AGI systems fundamentally depends on the interaction between personal memory and evolving memory.

\subsection{Memory as the Core Infrastructure for Human-Centric AGI}

From this perspective, the memory system supporting AGI must operate across three levels:

\begin{enumerate}

\item The \textbf{collective memory} derived from large-scale training data.

\item The \textbf{personal memory} accumulated through individual human–AI interactions.

\item The \textbf{evolving memory} produced through continuous reasoning and synthesis.

\end{enumerate}

A human-centric memory system must therefore support not only large-scale knowledge storage, but also mechanisms for filtering harmful knowledge, preserving personal ownership of experiences, and enabling knowledge co-evolution between humans and intelligent agents.

Consequently, memory should not be treated merely as a supporting component for agent execution. Instead, it forms a \textbf{core infrastructure layer} that governs how knowledge is accumulated, aligned, and evolved in human-centric AGI systems.
\section{Trends: From Agent-Centric Memory to Human-Centric Memory}

To understand the future role of memory in intelligent systems, it is useful to examine how the ownership and control of memory have evolved across technological paradigms.

Historically, digital memory systems have shifted between two dominant models:

\begin{itemize}
\item \textbf{Human-centric memory}, where individuals directly own and control their knowledge artifacts.
\item \textbf{Agent-centric memory}, where memory is managed internally by intelligent systems or service platforms.
\end{itemize}

Across the evolution of the internet and AI technologies, this relationship has undergone three major stages.

\subsection{Stage I: Human-Centric Memory on the Web}

In the early internet era, memory was primarily human-centric.

Individuals produced knowledge artifacts such as web pages, blog posts, documentation, and collaborative knowledge bases. These artifacts served as externalized representations of human memory and experience.

Users accessed and consumed this knowledge through search engines and hyperlinks. The internet functioned as a distributed repository of human-generated knowledge.

In this paradigm, memory ownership was naturally decentralized. Each individual or organization maintained control over the knowledge they published. Platforms primarily acted as indexing or hosting services rather than owners of the underlying knowledge.

From a systems perspective, memory was distributed across static documents, and knowledge integration remained a human-driven process.

\subsection{Stage II: Agent-Centric Memory in AI Systems}

The emergence of large language models fundamentally shifted memory control toward AI systems.

In this stage, knowledge is largely embedded within model parameters and centralized infrastructure operated by AI service providers. Users interact with intelligent agents through conversational interfaces, but the underlying memory structures are hidden inside model weights and proprietary databases.

Although AI systems significantly improve the efficiency of knowledge retrieval and synthesis, they introduce a strong form of \textbf{agent-centric memory}. Knowledge accumulation and memory evolution occur primarily within centralized training pipelines.

In practice, this means that:

\begin{itemize}
\item Memory is stored and managed by platforms.
\item Users have limited control over how their interactions contribute to system knowledge.
\item Knowledge evolution relies heavily on centralized retraining processes.
\end{itemize}

As a result, personal experiences and user-specific knowledge rarely persist beyond short-term interaction contexts. Long-term knowledge evolution remains costly and difficult to scale.

\subsection{Stage III: Human-Centric Memory as an Asset}

We are now entering a new stage in which memory returns to users and becomes a persistent digital asset.

Instead of being exclusively controlled by centralized AI systems, memory can be owned, accumulated, and managed by individuals while still being accessible to intelligent agents.

This paradigm shift introduces three key properties.

\paragraph{Memory in Hand}

First, individuals directly own and control their personal memories.

User experiences, preferences, and interaction histories are stored in user-managed memory spaces rather than exclusively within centralized platforms. These memories can persist across different agents and services, enabling long-term personalization and user sovereignty over digital knowledge.

\paragraph{Memory Group}

Second, memory systems support collaborative knowledge composition.

Individual memory spaces can selectively interact with others to form \textit{memory groups}. Through controlled sharing and access mechanisms, agents can utilize domain-specific knowledge stored in different personal memories.

This capability enables distributed expertise to be combined dynamically while maintaining privacy boundaries.

\paragraph{Collective Memory Evolution} Finally, memory systems support continuous knowledge evolution. 

Instead of relying solely on centralized model retraining, agents can synthesize new knowledge from accumulated experiences stored in memory. As described in Section~3, evolving memory is continuously updated through interactions between personal memory and existing evolving knowledge:

\begin{equation}
M_{e,t+1} = E(M_{p,t} \parallel M_{e,t})
\end{equation}

This mechanism allows knowledge to evolve organically over time, enabling intelligent systems to expand their capabilities through continuous interaction rather than periodic retraining.

\paragraph{State of current memory systems.}

Despite rapid progress in memory architectures for intelligent agents, existing systems only partially satisfy the three principles of the Memory-as-Asset paradigm: \emph{Memory in Hand}, \emph{Memory Group}, and \emph{Memory Evolution}. 
Different system designs emphasize different aspects of these principles, but none simultaneously achieves all three.

\begin{itemize}
    \item \textbf{Local memory systems} primarily support \emph{Memory in Hand (Human-Centric Ownership)}. By storing memory locally within a single device or agent runtime, they allow individuals to maintain strong control over their personal data and ensure privacy and portability. However, these systems typically operate in isolation. Memories remain bound to a single agent or application instance, forming \emph{memory islands} that cannot easily participate in collaborative knowledge sharing. Consequently, local systems struggle to support \emph{Memory Group} and provide limited capability for large-scale \emph{Memory Evolution}.

    \item \textbf{Centralized memory systems} partially enable \emph{Memory Group (Collaborative Knowledge Formation)}. By aggregating user memories within a shared platform infrastructure, they allow multiple agents and services to access common knowledge resources and support cross-application personalization. Nevertheless, memory ownership and control are tightly coupled with the platform provider, preventing individuals from maintaining true \emph{Memory in Hand}. Moreover, knowledge evolution is typically restricted within the boundaries of a single platform ecosystem.

    \item \textbf{Decentralized memory systems} move closer to supporting both \emph{Memory Group} and \emph{Memory Evolution (Continuous Knowledge Growth)}. Through distributed storage and interaction mechanisms, memories can be shared across agents and refined through collective experience. However, many existing decentralized designs still lack clear mechanisms for enforcing personal ownership and control, making it difficult to guarantee strong \emph{Memory in Hand}.
\end{itemize}

Taken together, current memory architectures reveal an inherent tension among these three properties. 
Existing systems typically achieve at most two of the three principles—\emph{Memory in Hand}, \emph{Memory Group}, and \emph{Memory Evolution}—but rarely all simultaneously. 
This limitation motivates the design of the \textit{Memory as Asset System}, which aims to integrate these principles within a unified human-centric memory architecture.

\subsection{Implications for Knowledge Evolution}

This shift fundamentally changes how knowledge evolves in intelligent systems.

Recall from Section~3 that the total knowledge space can be expressed as

\begin{equation}
K_{total} = K_c \cup K_p \cup K_e
\end{equation}

In agent-centric systems, knowledge growth is dominated by collective knowledge $K_c$ embedded in pretrained models. Personal knowledge $K_p$ and evolving knowledge $K_e$ remain limited due to the lack of persistent and user-controlled memory.

In contrast, human-centric memory infrastructures enable continuous growth of both $K_p$ and $K_e$ through persistent interaction and collaboration.

These observations suggest that future intelligent infrastructures must treat memory not merely as a storage component, but as a \textbf{first-class digital asset} that supports ownership, collaboration, and continuous knowledge evolution.
\section{Case Study: Memory as Asset System}

To operationalize the concept of decentralized agent memory, we present a system architecture called the \textit{Memory as Asset System (MAS)}. 
MAS treats memory not merely as an internal component of an agent, but as a persistent and tradable knowledge asset within a decentralized ecosystem.

The design of MAS follows the human-centric memory principle introduced in Section~3. 
Knowledge extracted from interactions can be divided into beneficial knowledge $K^{+}$ and harmful knowledge $K^{-}$. 
To ensure alignment with human values, only beneficial knowledge is allowed to be materialized into persistent memory:

\begin{equation}
M = G(K^{+})
\end{equation}

where $G(\cdot)$ denotes the memory materialization function that converts human-aligned knowledge into structured memory objects.

Under this formulation, memory becomes a structured asset derived from filtered knowledge. 
MAS provides the infrastructure necessary to create, store, evolve, and exchange such assets across agents.

The architecture of MAS consists of four major components:

\begin{itemize}

\item \textbf{Memory Object Model.}  
Defines the fundamental abstraction of memory as a structured knowledge asset with explicit ownership and access control.

\item \textbf{Personal Memory System.}  
Provides decentralized storage and management of memory objects owned by each agent.

\item \textbf{Memory Evolution Engine.}  
Implements the evolution process that continuously updates the agent’s memory space through interaction-driven learning.

\item \textbf{Memory Exchange Network.}  
Enables discovery, authentication, and exchange of memory assets across agents.

\end{itemize}

\begin{figure}[t]
\centering

\includegraphics[width=1\linewidth]{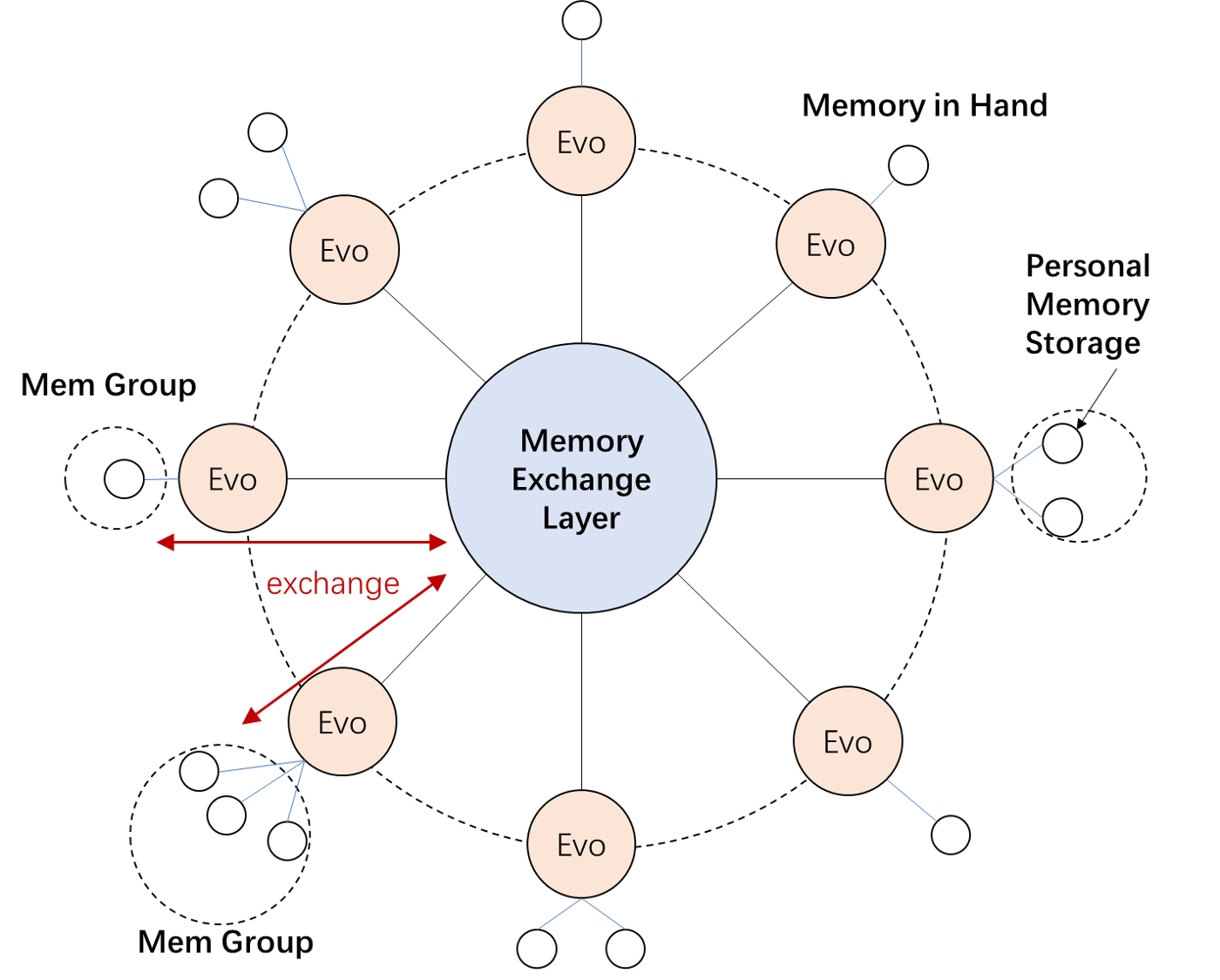}
\caption{
Architecture of the Memory as Asset System (MAS).
The outer ring represents the decentralized Personal Storage Layer, where each group maintains a personal memory space consisting of multiple personal agents. 
Each group connects to a local evolution node forming the Evolution Layer, responsible for updating the memory space through interaction-driven evolution. 
At the center lies the Memory Exchange network, which enables decentralized discovery and exchange of memory assets. 
The crossed symbols in the core represent the underlying network fabric supporting distributed memory exchange.
}

\label{fig:mas_architecture}
\end{figure}
Together, these components establish a decentralized infrastructure where memory can accumulate, evolve, and circulate as a knowledge asset.

\subsection{Design Goals}

MAS is designed around four principles that distinguish it from existing agent memory systems.

\textbf{Owner-Controlled Memory Ownership.}  
Memory assets are owned and governed by individuals rather than centralized service providers. 
Users maintain direct control over how their memories are stored, accessed, and managed across agents and services. This satisfies the Memory in Hand policy.

\textbf{Permission-Controlled Memory Assets.}  
Memory objects function as digital assets that can be selectively shared, licensed, or exchanged through explicit permission policies defined by their owners. This satisfies the Memory Group policy.

\textbf{Collective Memory Evolution.}  
Memory spaces evolve through knowledge exchange and interactions across agents and memory groups. 
By aggregating experiences from multiple participants, the system enables collective intelligence to emerge and continuously expand the knowledge boundary of agents.

\subsection{Memory Object Model}

The fundamental abstraction in MAS is the \textit{memory object}. 
A memory object represents a structured unit of human-centric knowledge derived from $K^{+}$ and stored in the memory space $M$.

Formally, a memory object is defined as

\begin{equation}
m = (c, e, o, \pi, v)
\end{equation}

where

\begin{itemize}

\item $c$ denotes the memory content derived from beneficial knowledge $K^{+}$,
\item $e$ represents contextual metadata,
\item $o$ identifies the memory owner,
\item $\pi$ specifies the access permission policy,
\item $v$ records the version or evolution state of the memory.

\end{itemize}

\subsection{Personal Memory System}

Each agent maintains a \textit{Personal Memory System} that stores its private memory space $M_t$.

This component is responsible for organizing memory objects generated from interactions and acquired from external sources. 
The personal memory system transforms filtered knowledge into persistent memory assets through the materialization function:

\begin{equation}
M_t = G(K_t^{+})
\end{equation}

The personal memory system provides three essential capabilities.

\textbf{Ownership Control.}  
Each memory object is owned by the agent that generated or acquired it.

\textbf{Structured Storage.}  
Memory objects are stored in structured repositories that support efficient retrieval and long-term persistence.

\textbf{Privacy Protection.}  
Sensitive knowledge may remain private within the owner's memory space.

By separating personal memory from the global ecosystem, MAS allows agents to maintain full sovereignty over their knowledge assets.

\subsection{Memory Evolution Engine}

The Memory Evolution Engine is deployed on the user or agent side and is responsible for continuously evolving the agent’s memory space.

As defined in Section~3, memory consists of two components: private memory $M_p$ and evolving memory $M_e$. 
The evolution process updates the evolving memory state according to

\begin{equation}
M_{e,t+1} = E(M_{p,t} \Vert M_{e,t})
\end{equation}

where $E(\cdot)$ denotes the memory evolution operator and $\Vert$ represents memory composition.

This formulation captures the idea that new memory is generated by combining stable personal knowledge with accumulated evolving experiences.

In practice, the evolution engine performs several transformations:

\textbf{Experience Recording.}  
New interaction outcomes are recorded as memory objects.

\textbf{Memory Consolidation.}  
Related memory objects may be merged to form more coherent knowledge representations.

\textbf{Knowledge Abstraction.}  
Higher-level knowledge structures may be synthesized from multiple experiences.

\textbf{Human-Centric Filtering.}  
The system ensures that only beneficial knowledge contributes to memory evolution.

Through these transformations, the engine continuously updates the memory space and expands the agent's knowledge capacity.

\subsection{Memory Exchange Network}

To support large-scale knowledge collaboration, MAS introduces a decentralized \textit{Memory Exchange Network}.

In this model, memory objects function as tradable digital knowledge assets. 
Each memory object has a clearly defined owner and permission policy, allowing it to be exchanged under controlled conditions.

The exchange layer supports several core operations.

\textbf{Discovery.}  
Agents can search for relevant memory objects across the decentralized ecosystem.

\textbf{Authentication and Authorization.}  
Access to memory assets is governed by ownership and permission policies.

\textbf{Asset Exchange.}  
Memory objects may be shared, licensed, or traded between agents.

When an agent acquires an external memory object $m_j$, the object can be incorporated into the evolving memory space:

\begin{equation}
M_{e,t+1} = E(M_{p,t} \Vert (M_{e,t} \cup m_j))
\end{equation}

This mechanism enables agents to extend their knowledge beyond locally generated experiences while preserving ownership and permission control.

Through decentralized exchange, the MAS infrastructure transforms isolated agent memories into a collaborative knowledge ecosystem.
\section{Discussion}

\paragraph{Deployment Flexibility and Practical Considerations.}
Although the Memory-as-Asset System (MAS) emphasizes human-centric ownership of memory, this does not necessarily imply that all memory storage must be physically localized on user devices. 
In practical deployments, personal memory spaces may be hosted either locally or by service providers, depending on performance, cost, and usability considerations. 
However, MAS requires that each individual maintains a logically private memory space with strong ownership guarantees. 
When memory is hosted by external service providers, the stored data should be protected through encryption mechanisms (e.g., public-key encryption), ensuring that memory assets remain accessible only to their rightful owners.

\paragraph{Flexible Placement of Memory Evolution.}
The memory evolution process is similarly flexible in its deployment. 
The evolution function $E(\cdot)$ may execute locally on personal devices or remotely on more powerful infrastructure depending on the computational requirements. 
Local evolution preserves stronger privacy guarantees, while remote evolution can leverage scalable computing resources to accelerate knowledge synthesis. 
MAS therefore supports hybrid deployment models in which memory evolution can occur across both personal and shared computational environments.

\paragraph{Permission-Controlled Memory Sharing.}
Finally, while decentralized memory management can improve resilience and reduce platform dependence, it is not strictly required for the MAS paradigm. 
What remains essential, however, is the ability to enforce fine-grained permission control over memory sharing. 
Memory assets must support explicit access policies that regulate how memories are shared across agents, users, and memory groups. 
Such permission-controlled sharing mechanisms ensure that collaborative knowledge formation and collective memory evolution can occur without compromising personal ownership and privacy.
\section{Conclusion}

In this paper, we argue that memory will become a foundational infrastructure for the next generation of intelligent agents. While recent advances in large language models have significantly improved reasoning and interaction capabilities, most existing agent systems still rely on static model knowledge or short-term context, limiting their ability to accumulate and evolve knowledge over time.

We introduce the concept of \textit{Memory as Asset}, a new paradigm that elevates memory from a local component of individual agents to a shared and evolvable knowledge infrastructure. 
By treating memory as a first-class asset, we enable decentralized memory storage, cross-agent knowledge exchange, and continuous memory evolution. 
This paradigm allows intelligent agents to move beyond isolated reasoning toward collaborative knowledge accumulation across a global memory ecosystem.

We further discuss the architectural foundations of \textit{Memory as Asset System} (MAS), including the abstraction of memory objects, decentralized memory infrastructure, memory exchange mechanisms, and memory-driven knowledge evolution. 
Together, these concepts outline a scalable path to building open memory ecosystems where knowledge can be stored, shared, and refined across agents and organizations.

We believe that transforming memory into an open, decentralized service will play a critical role in enabling long-term learning and collective intelligence. As intelligent systems increasingly interact with complex environments and diverse users, \textit{Memory as Asset} offers a promising approach to building persistent knowledge infrastructures that support the continuous growth of machine intelligence.

\bibliographystyle{unsrtnat}
\bibliography{amaas}

\end{document}